\documentstyle[lingmacr,epsf,psfig,rotating,twoside]{article}

\author{{\em Melanie Siegel}\\ \\ {\normalsize Department of Computational
Linguistics}\\{\normalsize University of the Saarland}\\{\normalsize Postfach 151150}\\{\normalsize D-66041
Saarbr\"ucken, Germany}\\{\normalsize {\tt siegel@dfki.de}}} 

\title{{\bf THE SYNTACTIC PROCESSING OF PARTICLES IN JAPANESE SPOKEN LANGUAGE}} 

\date{}

\begin{document}

\newcommand{\VM}{\mbox{{\sf Verb}{\textsf {\textsl {mobil\/}}}}}

\newcommand{\JPChar}[1]{
\renewcommand{\arraystretch}{0.5}
 \begin{tabular}{l}
    #1
  \end{tabular}
}

\newtheorem{exmpl}{}

\bibliographystyle{plain}

\maketitle
\thispagestyle{empty}

\vspace{-1.2cm}
\begin{centering}\section*{Abstract}\end{centering}

Particles fullfill several distinct central roles in the Japanese
language. They can mark arguments as well as adjuncts, can be
functional or have semantic funtions. There is, however, no
straightforward matching from particles to functions, as, e.g., {\em
ga} can mark the subject, the object or an adjunct of a
sentence. Particles can cooccur. Verbal arguments that
could be identified by particles can be eliminated in the Japanese
sentence. And finally, in spoken language particles are often omitted.
A proper treatment of particles is thus necessary to make an analysis
of Japanese sentences possible. Our treatment is based on an empirical
investigation of 800 dialogues. We set up a type hierarchy of particles
motivated by their subcategorizational and modificational
behaviour. This type hierarchy is part of the Japanese syntax in VERBMOBIL.

\vspace{7mm}
\begin{centering}\section{Introduction}\end{centering}

The treatment of particles is essential for the processing of the
Japanese language for two reasons. The first reason is that these are
the words that occur most frequently. The second reason is that
particles have various central functions in the Japanese syntax: case
particles mark subcategorized verbal arguments, postpositions mark
adjuncts and have semantic attributes, topic particles mark topicalized
phrases and {\em no} marks an attributive nominal adjunct.  Their
treatment is difficult for three reasons: 1) despite their central
position in Japanese syntax the omission of particles occurs quite
often in spoken language. 2) One particle can fulfill more than one
function.  3) Particles can cooccur, but not in an arbitrary way.

In order to set up a grammar that accounts for a larger amount of
spoken language, a comprehensive investigation of Japanese particles is
thus necessary.  Such a comprehensive investigation of Japanese
particles was missing up to now. Two kinds of solutions have previously
been proposed: (1) the particles are divided into case particles and
postpositions. The latter build the heads of their phrases, while the
former do not (cf. \cite{Miyagawa86}, \cite{Tsujimura96}). (2) All
kinds of particles build the head of their phrases and have the same
lexical structure (cf. \cite{Gunji87}). Both kinds of analyses lead to
problems: if postpositions are heads, while case particles are
nonheads, a sufficient treatment of those cases where two or three
particles occur sequentially is not possible, as we will show. If on
the other hand there is no distinction of particles, it is not possible
to encode their different behaviour in subcategorization and
modification.
We carried out an empirical investigation of cooccurrences of particles
in Japanese spoken language. As a result, we could set up restrictions
for 25 particles.  We show that the problem is essentially based at the
lexical level. Instead of assuming different phrase structure rules we
state a type hierarchy of Japanese particles. This makes a uniform
treatment of phrase structure as well as a differentiation of
subcategorization patterns possible. We therefore adopt the `all-head'
analysis, but extend it by a type hierarchy in order to be able to
differentiate between the particles.  Our analysis is based on 800
Japanese dialogues of the VERBMOBIL data concerning appointment scheduling.

\begin{centering}\section{The Type Hierarchy of Japanese Particles}\end{centering}

Japanese noun phrases can be modified by more than one particle at a
time. There are many examples in our data where two or three particles
occur sequentially. On the one hand, this phenomenon must be accounted
for in order to attain a correct processing of the data. On the other
hand, the discrimination of particles is motivated by their
modificational and subcategorizational behaviour. We carried out an
empirical analysis, based on our dialogue
data. Table~\ref{fig:cooc-data} shows the frequency of cooccurrence of
two particles in the dialogue data. There is a tendency to avoid the
cooccurrence of particles with the same phonology, even if it is
possible in principal in some cases. The reason is obvious: such
sentences are difficult to understand.

\noindent
\begin {table}[ht]
\renewcommand{\arraystretch}{0.5}
\begin{small}
\begin{tabular}{|l|l|l|l|l|l|l|l|l|l|l|l|l|l|}
left$\downarrow$/right$\rightarrow$&ga&wo&ni&de&e&kara&made&no&wa&mo&naNka&to\\
\hline
ga&0&0&0&0&0&0&0&0&0&0&0&0\\
\hline
wo&0&0&0&0&0&0&0&0&0&0&0&3\\
\hline
ni&0&0&0&19&0&0&0&0&137&49&0&15\\
\hline
de&2&0&0&0&0&0&0&14&158&241&0&30\\
\hline
e&0&0&0&1&0&0&0&4&0&0&0&0\\
\hline
kara&23&0&30&81&0&0&0&34&69&12&0&123\\
\hline
made&17&1&66&32&0&0&0&40&63&1&0&79\\
\hline
no&64&9&1&2249&0&0&0&0&287&11&0&4\\
\hline
wa&0&0&0&2&0&0&0&0&0&0&1&3\\
\hline
mo&0&0&0&0&0&0&0&0&0&0&0&0\\
\hline
naNka&3&0&0&1&0&0&0&0&30&0&0&0\\
\hline
to&0&3&0&1&0&0&0&14&17&58&0&0\\
\hline
toshite&0&0&0&0&0&0&0&0&36&15&0&0\\
\hline
toshimashite&0&0&0&0&0&0&0&0&15&0&0&0\\

\hline
\end{tabular}
\caption{Cooccurrence of 2 Particles in the 800 Dialogues}
\label{fig:cooc-data}
\end{small}
\end{table}

\cite{Kuno73} treats {\em wa, ga, wo, ni, de, to, made, kara} and {\em
ya} as `particles'. They are divided into those that are in the deep
structure and those that are introduced through transformations. An
example for the former is {\em kara}, examples for the latter are {\em
ga}(SBJ), {\em wo}(OBJ), {\em ga}(OBJ) and {\em ni}(OBJ2).
\cite{Gunji87} assigns all particles the part-of-speech P. Examples are
{\em ga, wo, ni, no, de, e, kara} and {\em made}. All particles are
heads of their phrases. Verbal arguments get a grammatical relation [GR
OBJ/SBJ]. In \cite{Gunji91} the part-of-speech class P contains only
{\em ga, wo} and {\em ni}.
\cite{Tsujimura96} defines postpositions and case particles such that
postpositions are the Japanese counterpart of prepositions in English
and cannot stand independently, while case particles assign case and
can follow postpositions. Her case particles include {\em ga, wo, ni,
no} and {\em wa}.
\cite{Nightingale96} divides case markers ({\em ga, wo, ni} and {\em wa})
from copula forms ({\em ni, de, na} and {\em no}). He argues that {\em
ni, de, na} and {\em no} are the infinitive, gerund and adnominal
forms of the copula.

\begin{figure}[t]
\psfig{figure=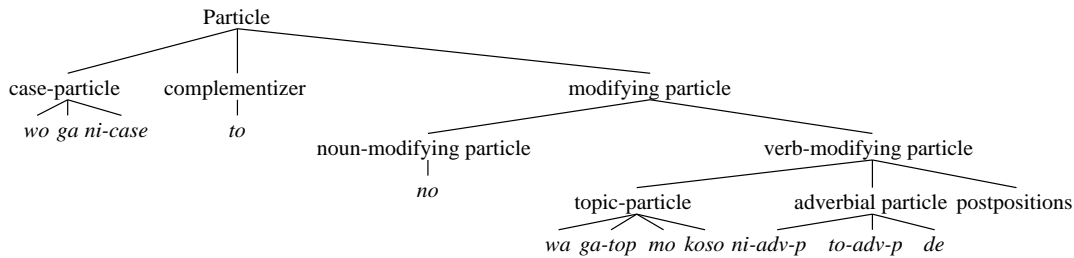,width=15cm}
\vspace{-1.2cm}
\caption{{\small Type Hierarchy of Japanese Particles. 
Postpositions include {\em e, naNka, sonota, tomo, kara, made, soshite,
nado, bakari, igai, yori, toshite, toshimashite, nitsuite, nikaNshite}
and {\em nikakete}.}}\label{figure:types}
\end{figure}

In the class of particles, we include case particles,
complementizers, modifying particles and conjunctional particles. We
thus assume a common class of the several kinds of particles introduced
by the other authors. But they are further divided into subclasses, as
can be seen in figure~\ref{figure:types}. We assume not only a
differentiation between case particles and postpositions, but a finer
graded distinction that includes different kinds of particles not
mentioned by the other authors. {\em de} is assumed to be a particle
and not a copula, as \cite{Nightingale96} proposes. It belongs to the class of
adverbial particles.  One major motivation for the
type hierarchy is the observation we made of the cooccurrence of
particles.
Case particles ({\em ga, wo, ni}) are those that attach to verbal
arguments. A complementizer marks
complement sentences. Modifying particles attach to adjuncts. They are
further divided into noun-modifying particles and verb-modifying
particles. Verb modifying particles can be topic particles, adverbial
particles, or postpositions. Some particles can have more than one function, as for
example {\em ni} has the function of a case particle and an adverbial
particle.
Figure~\ref{figure:types} shows the type hierarchy of Japanese
particles. The next sections examine the individual types of particles.

\subsection{Case Particles}

There is no number nor gender agreement between noun phrase and
verb. The verbs assign case to the noun phrases. This is marked by the
case particles. Therefore these have a syntactic function, but not a
semantic one. Unlike in English, the grammatical functions cannot
be assigned through positions in the sentence or c-command-relations,
since Japanese exhibits no fixed word position for verbal arguments. The
assignment of the grammatical function is not expressed by the case
particle alone but only in connection with the verbal valency. There
are verbs that require {\em ga}-marked objects, while in most cases the
{\em ga}-marked argument is the subject:

\vspace{-0.3cm}
\enumsentence{
\label{eq:3}

\JPChar{nantoka\\somehow}
\JPChar{yotei\\time}
\JPChar{ga\\GA}
\JPChar{toreru\\can take}
\JPChar{N desu\\COP}
\JPChar{ga\\SAP}\\
(Somehow (I) can find some time.)}

\vspace{-0.3cm}
Japanese is described as a head-final language. \cite{Gunji87}
therefore assumes only one phrase structure rule: {\bf M}{\small other}
$\longrightarrow$ {\bf D}{\small aughter} {\bf H}{\small ead}. However, research
literature questions whether this also applies to nominal phrases and
their case particles.  \cite{HPSG}:45 assume Japanese case particles to
be markers.

On the one hand, there are several reasons to  distinguish case particles
and modifying particles. On the other hand, I doubt
whether it is reasonable to assume different {\em phrase structures} for
NP+case particle and NP+modifying particle. 
The phrase-structural distinction of case particles and postpositions
leads to problems, when more than one particle occur. The following
example comes from the {\VM} corpus:

\vspace{-0.3cm}
\enumsentence{
\label{eq:7}
\JPChar{naNji\\what time}
\JPChar{kara\\from}
\JPChar{ga\\GA}
\JPChar{yoroshii\\good}
\JPChar{desu\\COP}
\JPChar{ka\\QUE}\\
(At what time would you like to start?)

}

\begin{figure}[b]
\epsffile{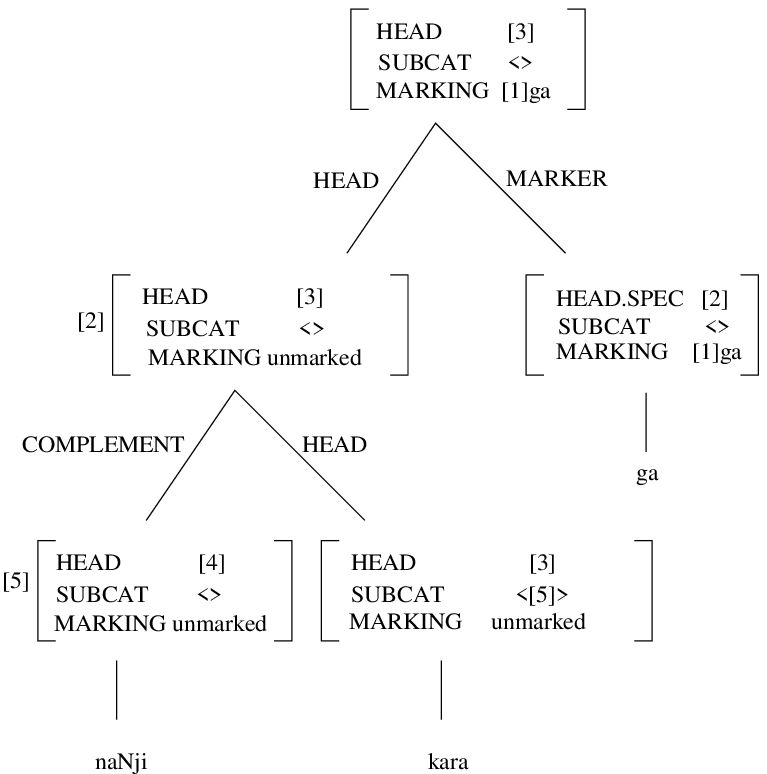}
\vspace{-0.5cm}
\caption{{\em NaNji kara ga} with Head-Marker Structure}\label{figure:nanjikaraga}
\vspace{-0.5cm}
\end{figure}

\vspace{-0.3cm}
If one now assumes that the modifying particle {\em kara}  is head of
{\em naNji} as well as of the case particle {\em ga}, the result for {\em naNji kara ga} with the head-marker
structure  described in \cite{HPSG}\footnote{The
Marking Principle says: {\em In a headed phrase, the MARKING value is
token-identical with that of the MARKER-DAUGHTER if any, and with that
of the HEAD-DAUGHTER otherwise\cite{HPSG}.}} would be as shown in
figure~\ref{figure:nanjikaraga}. 
The case particle {\em ga} would have to allow nouns and modifying
particles in SPEC. The latter are however normally adjuncts that modify
verbal projections.
Therefore the head of {\em kara} entails the information that it can
modify a verb. This information is inherited to the head of the whole
phrase by the Head-Feature Principle as is to be seen in the tree
above.  As a result, this is also admitted as an adjunct to a verb,
which leads to wrong analyses for sentences like the following one:

\vspace{-0.3cm}
\enumsentence{

\label{eq:27}
\JPChar{*naNji\\what time}
\JPChar{kara\\from}
\JPChar{ga\\GA}
\JPChar{sochira\\you}
\JPChar{ga\\GA}
\JPChar{jikaN\\time}
\JPChar{ga\\GA}
\JPChar{toremasu\\can take}
\JPChar{ka\\QUE}

}

\vspace{-0.3cm}
If, on the other hand, case particles and topic markers are heads, one
receives a consistent and correct processing of this kind of example
too. This is because the head information [MOD none] is given from the
particle {\em ga} to the head of the phrase {\em naNji kara ga}. Thus
this phrase is not admitted as an adjunct.

Instead of assuming different phrase structure rules, a distinction of
the kinds of particles can be based on lexical types. HPSG offers the
possibility to define a common type and to set up specifications for
the different types of particles.
We assume Japanese to be head-final in this respect. All kinds
of particles are analysed as heads of their phrases.
The relation between case particle and nominal phrase is a
`Complement-Head' relation. The complement is obligatory and
adjacent\footnote{Obligatory Japanese arguments are always adjacent,
and vice versa.}.
Normally the case particle {\em ga} marks the subject, the case
particle {\em wo} the direct object and the case particle {\em ni} the
indirect object. There are, however, many exceptions. We therefore use
predicate-argument-structures instead of a direct assignment of
grammatical functions by the particles (and possibly
transformations). The valency information of the Japanese verbs does
not only contain the syntactic category and the semantic restrictions
of the subcategorized arguments, but also the case particles they must
be annotated with\footnote{\cite{Ono94} investigates the particles
{\em ni}, {\em ga} and {\em wo} and also states that grammatical
functions must be clearly distinguished from surface cases}.

In most cases the {\em ga}-marked noun phrase is the subject of the
sentence.
However, this is not always the case. Notably stative verbs
subcategorize for {\em ga}-marked objects. An example is the stative
verb {\em dekimasu}\footnote{see \cite{Kuno73} for a semantic
classification of verbs that take {\em ga}-objects}:

\vspace{-0.3cm}
\enumsentence{

\label{eq:29}

\JPChar{kanojo\\she}
\JPChar{ga\\GA}
\JPChar{oyogi\\swimming}
\JPChar{ga\\GA}
\JPChar{dekimasu\\can}\\
(She can swim.)
}

\vspace{-0.3cm}
These and other cases are sometimes called `double-subject
constructions' in the literature. But these {\em ga}-marked noun
phrases do not behave like subjects. They are neither subject to
restrictions on subject honorification nor subject to reflexive binding
by the subject. This can be shown by the following example:

\vspace{-0.3cm}
\enumsentence{

\label{eq:25}

\JPChar{gogo\\afternoon}
\JPChar{no\\NO}
\JPChar{hou\\side}
\JPChar{ga\\GA}
\JPChar{yukkuri\\at ease}
\JPChar{{\em hanashi}\\talking}
\JPChar{ga\\GA}
\JPChar{dekimasu\\can}
\JPChar{ne\\SAP}\\
(We can talk at ease in the afternoon.)
}

\vspace{-0.3cm}
{\em hanashi} does not meet the semantic restriction [+animate]
stated by the verb {\em dekimasu} for its subject. 
There are even {\em ga}-marked adjuncts.
\cite{Kuroda92} assumes these `double-subject constructions' to
be derived from genitive relations. 
But this analysis seems not to be true for example~\ref{eq:25}), because
the following sentence is wrong:

\vspace{-0.3cm}
\enumsentence{

\label{eq:83}

\JPChar{*gogo\\afternoon}
\JPChar{no\\NO}
\JPChar{hou\\side}
\JPChar{no\\NO}
\JPChar{yukkuri\\at ease}
\JPChar{hanashi\\talk}
\JPChar{ga\\GA}
\JPChar{dekimasu\\can}
\JPChar{ne\\SAP}
}

\vspace{-0.3cm}

The case particle {\em wo} normally marks the direct object of the
sentence.
In contrast to {\em ga}, no two phrases in one clause may be marked by {\em
wo}. This restriction is called `double-wo constraint' in research
literature (see, for example, \cite{Tsujimura96}:249ff.). 
Object positions with {\em wo}-marking as well as subject positions with
{\em ga}-marking can be
saturated only once. There are neither double subjects nor double
objects. This restriction is also valid for
indirect objects. Arguments found must be assigned a saturated status in the
subcategorization frame, so that they cannot be saturated again (as
in English). The verbs subcategorize for at most one
subject, object and indirect object. Only one of these arguments may be
marked by {\em wo}, while a subject and an object may both be marked by
{\em ga}. These attributes are determined by the verbal valency.
The {\em wo}-marked argument is not required to be adjacent to the verb. It is
possible to reverse NP-{\em ga} and NP-{\em wo} as well as to insert
adjuncts between the arguments and the verb.

The particle {\em ni} can have the function of a case particle as well
as that of an adjunct particle modifying the
predicate. \cite{Sadakane/Koizumi95} also identify homophoneous {\em
ni} that can mark adjuncts or complements. They use the notion of
`affectedness' to distinguish them. This is however not useful in our
domain. \cite{Ono94} suggest testing the possibility of
passivization. 
Some verbs subcategorize for a {\em ni}-marked object, as for example
{\em naru}:

\vspace{-0.3cm}
\enumsentence{

\label{eq:51}

\JPChar{raigetsu\\next month}
\JPChar{ni\\NI}
\JPChar{naru\\become}
\JPChar{N desu\\COP}
\JPChar{ga\\SAP}\\
(It will be next month.)
}

\vspace{-0.3cm}
{\em ni}-marked objects cannot occur twice in the same clause, just as
{\em ga}-marked subjects and {\em wo}-marked objects. The `double-wo
constraint' is neither a specific Japanese restriction nor a specific
peculiarity of the Japanese direct object. It is based on the wrong
assumption that grammatical functions are assigned by case
particles. There are a lot of examples with double NP-{\em ni}, but
these are adjuncts.

The lexical entries of case particles get a case entry in the
HEAD. Possible values are {\em ga}, {\em wo}, {\em ni} and {\em to}. They are neither adjuncts nor specifiers and thus get the entries
[MOD none] and [SPEC none]. They subcategorize for an adjacent
object. This can be a noun, a postposition or an adverbial particle\footnote{A fundamental difference between Japanese grammar and English
grammar is the fact that verbal arguments can be optional. For example,
subjects and objects that refer to the speaker are omitted in most
cases in spoken language. The verbal arguments can freely scramble.
Additionally, there exist adjacent verbal arguments.
To account for this, our subcategorization contains the attributes SAT
and VAL. In SAT it is noted, whether a verbal argument is already
saturated (such that it cannot be saturated again), optional or
adjacent. VAL contains the agreement information
for the verbal argument. 
Adjacency must be checked in every rule that combines heads and
arguments or adjuncts.}.

\begin{figure}
\psfig{figure=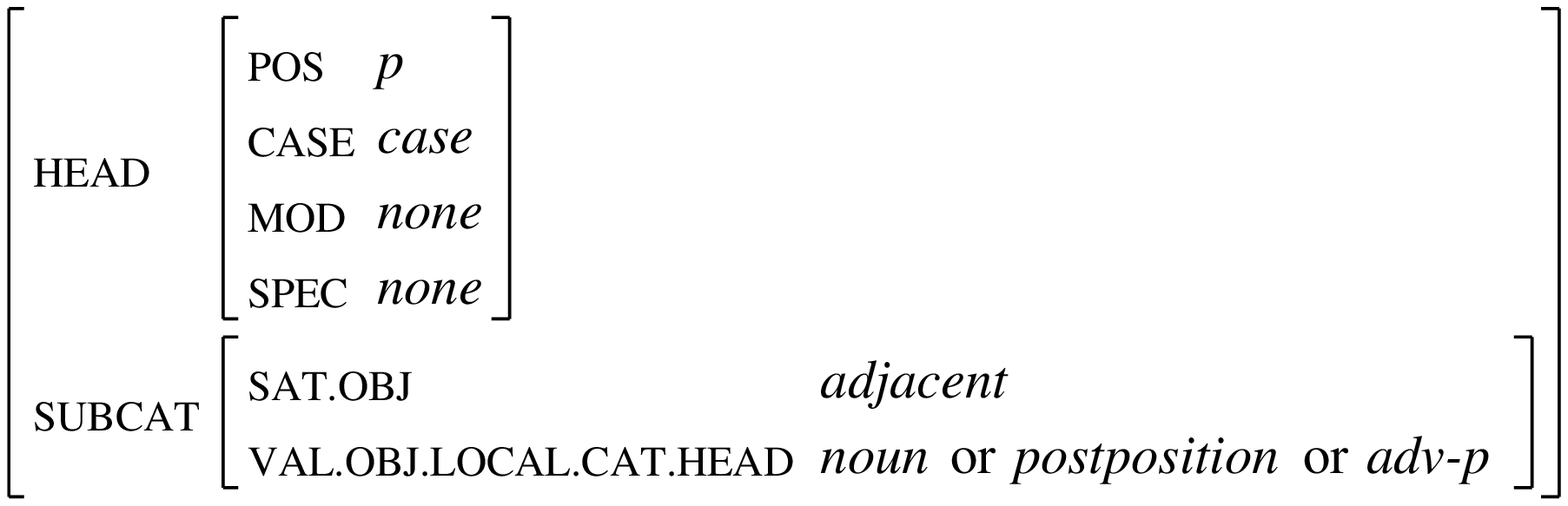,width=10cm}
\vspace{-1.7cm}
\caption{Head and Subcat of Case Particles}
\label{figure: Case-Particles-AVM}
\end{figure}

\subsection{The Complementizer {\em to}}

{\em to} marks adjacent complement sentences that are
subcategorized for by verbs like {\em omou}, {\em iu} or {\em kaku}.

\vspace{-0.3cm}
\enumsentence{

\label{eq:106}

\JPChar{sochira\\you}
\JPChar{ni\\NI}
\JPChar{ukagaitai\\visit}
\JPChar{to\\TO}
\JPChar{omoimasu\\think}
\JPChar{node\\SAP}\\
(I would like to visit you.)
}

\vspace{-0.3cm} Some verbs subcategorize for a {\em to} marked
object. This object can be optional or obligatory with verbs like {\em
kuraberu}.

\enumsentence{

\label{eq:110}

\JPChar{kono\\that}
\JPChar{hi\\day}
\JPChar{mo\\too}
\JPChar{chotto\\somewhat}
\JPChar{hito\\people}
\JPChar{to\\TO}
\JPChar{au\\meet}
\JPChar{yotei\\plan}
\JPChar{ga\\GA}
\JPChar{gozaimasu\\exist}\\
(That day too, there is a plan to meet some people.)
}

\vspace{-0.3cm} {\em to} in these  cases is categorized as a
complementizer. Another possibility is that {\em to} marks an
adjunct to a predicate, which qualifies {\em to} as a verb modifying
particle:

\vspace{-0.3cm}
\enumsentence{

\label{eq:111}

\JPChar{shimizu\\Shimizu}
\JPChar{seNsei\\Prof.}
\JPChar{to\\TO}
\JPChar{teNjikai\\exhibition}
\JPChar{wo\\WO}
\JPChar{go-issho\\together}
\JPChar{sasete\\do}
\JPChar{itadaku\\HON}\\
(I would like to organize an exhibition with Prof. Shimizu.)
}

\vspace{-0.3cm} Finally, the complementizer {\em to} can be an NP
conjunction (which will not be considered at the moment, see
\cite{Kuno73}).  The complementizer gets a case entry, because its head
is a subtype of {\em case-particle-head}. It subcategorizes for a noun,
a verb, an utterance, an adverbial particle or a postposition.

\subsection{Modifying Particles}

An essential problem is to find criteria for the distinction of case
particles and modifying particles. On the semantic level they can be
distinguished in that modifying particles introduce semantics, while
case particles have a functional meaning. According to this, the
particle {\em no} is a modifying one, because it introduces attributive
meaning, as opposed to (\cite{Tsujimura96}:134), who classifies it as a
case particle. Another distinctive criterion that is introduced by
\cite{Tsujimura96}:135 says that modifying particles\footnote{He calls
them `postpositions'.} are obligatory in spoken language, while case
particles can be omitted.  Case particles are indeed suppressed more
often, but there are also cases of suppressed modifying
particles. These occur mainly in temporal expressions in our dialogue
data:

\vspace{-0.3cm}
\enumsentence{

\label{eq:57}

\JPChar{soredewa\\then}
\JPChar{juuyokka\\14th}
\JPChar{no\\NO}
\JPChar{gogo\\afternoon}
\JPChar{\O\\\O}
\JPChar{niji\\2 o'clock}
\JPChar{\O\\\O}
\JPChar{robii\\lobby}
\JPChar{no\\NO}
\JPChar{hou\\side}
\JPChar{de\\DE}
\JPChar{o machi\\HON-wait}
\JPChar{shite\\do}
\JPChar{orimasu\\AUX-HON}\\
(I will then wait in the lobby  at 2 o'clock on the 14th.)
}

\vspace{-0.3cm}
Finally \cite{Tsujimura96} gives the criterion that case particles can follow
modifying particles while modifying particles cannot follow case
particles. This criterion in particular implies that a finer
distinction is necessary, as we have shown that it is not that
easy. This can be realized with HPSG types. According to this
criterion, {\em no} behaves like a modifying particle, while according
to the criterion on meaning, it behaves like a case particle. Our first
distinction is thus a functional one: modifying particles differ from
case particles in that their marked entities are not subcategorized for
by the verb. Case particles get the head information [CASE case] that
controls agreement between verbs and their arguments.  Modifying
particles do not get this entry. They get the information in MOD that
they can become adjuncts to verbs (verb modifying particles) or nouns
(the noun modifying particle {\em no}) and semantic information. They
subcategorize for a noun, as all particles do. The modifying particles
share the following features in their lexical entries.

\begin{figure}
\psfig{figure=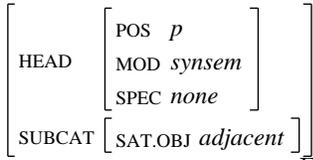,width=5cm}
\vspace{-1.8cm}
\caption{Head and Subcat of Modifying Particles}\label{figure:mod-p-avm}
\end{figure}

\vspace{-0.3cm}
\subsubsection{Verb Modifying Particles}

The verb modifying particles specify the modification of the verb in
MOD.
The postpositions modify a (nonauxiliary) verb as an adjunct and
subcategorize for a nominal object.
\cite{Nightingale96} treats {\em ni} and {\em de} as the infinitive and
the gerund form of the copula. {\em ni} is similar to the infinitive
form to the extend that it can take an adverb as its argument
({\em gogo wa furii ni nat-te i-masu} -- afternoon - WA - free -
become). But the infinitive is clearly distinct from the
characteristics of {\em ni}, that cannot be used with {\em N desu},
cannot mark a relative sentence ({\em *John ga furii ni koto}) and
cannot be marked with the complementizer {\em to} ({\em *John ga furii
ni to omou}).
The adjunctive form `de' has both qualities of a gerundive copula and
qualities of a particle. But there is some data that shows different
behaviour of {\em de} and other gerundives. Firstly, it concerns the
cooccurrence possibilities of {\em de} and other particles, compared to
gerundive forms and particles:

\begin{tabular}{llll}
de wa - V-te wa & de mo - V-te mo & de no - V-te no & de ni - *V-te ni\\
de ga - *V-te ga & de wo - *V-te wo & de de - *V-te de & \\
\end{tabular}

\noindent Secondly, a gerund may modify auxiliaries,
e.g. {\em shite kudasai, shite orimasu}, but {\em de} may not. 
Additionally there is something which distinguishes {\em de} of a copula: it may
not subcategorize for a subject. 
A word that is an adjunct to verbs, subcategorizes for an unmarked noun
or a postpositional phrase and is subcategorized for by several
particles (see above) fits well into our description of a verb
modifying particle.
The adverbial particles {\em ni}, {\em de} and {\em to} subcategorize for a noun
or a postposition.
As already described, {\em to} behaves like an adverbial particle,
too.

\vspace{-0.3cm}
\subsubsection{The Noun Modifying Particle {\em NO}} 

{\em no} is a particle that modifies nominal phrases.  This is an
attributive modification and has a wide range of meanings.\footnote{See
also \cite{Tsuda/Harada96}}
\cite{Tsujimura96}:134ff. assigns {\em no} to the class of case
particles. However, the criteria she sets up to distinguish between case
particles and postpositions do not apply to this classification of {\em
no}: firstly, Tsujimura's postpositions have their own semantic
meaning. Case particles have a functional meaning. {\em no} however has
a semantic, namely attributive meaning. Secondly, Tsujimura's
postpositions are obligatory in spoken language, case particles are
optional. {\em no} is as obligatory as {\em kara} and {\em
made}. Finally, Case particles can - as Tsujimura states - follow
postpositions, but postpositions cannot follow case
particles. According to this criterion, {\em no} behaves like a case
particle.
{\em no} combines qualities of case particles with those of modifying
particles (which Tsujimura calls `postpositions'). This means that a
special treatment of this particle is necessary.  The particle {\em no}
subcategorizes for a noun, as the other particles do. It also modifies
a noun. This separates it from the other modifying particles. The 
particle {\em no} modifies a noun phrase and occurs after a noun or a
verb modifying particle.

\vspace{-0.3cm}
\subsubsection{Particles of Topicalization}

The topic particle {\em wa} can mark arguments as well as adjuncts. In
the case of argument marking it replaces the case particle. In the case
of adjunct marking it can replace the verb modifying particle or it can
occur after it.  On the syntactic level, it has to be decided, whether
the topic particle marks an argument or an adjunct, when it occurs
without a verb modifying particle. This is difficult because of the
optionality of verbal arguments in Japanese. If it marks an argument,
it has to be decided which grammatical function this argument has. This
problem can often not be solved on the purely syntactic level. Semantic
restrictions for verbal arguments are necessary:

\vspace{-0.3cm}
 \enumsentence{

\label{eq:81}

\JPChar{basho\\place}
\JPChar{no\\NO}
\JPChar{hou\\side}
\JPChar{wa\\WA}
\JPChar{dou\\how}
\JPChar{shimashou\\shall do}
\JPChar{ka\\QUE}\\
(How shall we resolve the problem of the place?)
}

\vspace{-0.3cm}
Subject and object of the verb {\em shimashou} are suppressed in this
example. The sentence can be interpreted as having a topic adjunct, but no
surface subject and object, when using semantic restrictions for the
subject ({\em agentive}) and the object ({\em situation}).

\begin{figure}
\psfig{figure=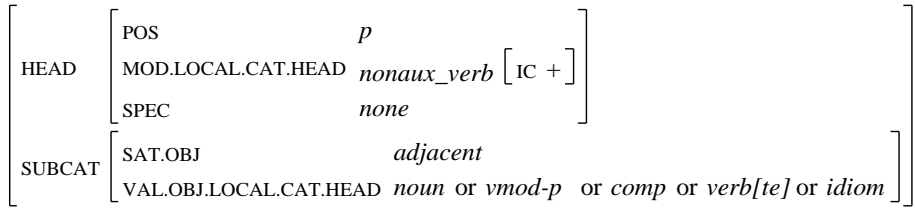,width=13cm}
\vspace{-1.8cm}
\caption{Topic Particle AVM}\label{figure:topic-p-avm}
\end{figure}

\cite{Gunji91} analyses Japanese topicalization with a trace that
introduces a value in SLASH and the `Binding Feature Principle' that
unifies the value of SLASH with a {\em wa}-marked element\footnote{The
Binding Feature Principle says: {\em The value of a binding feature of
the mother is identical to the union of the values of the binding
feature of the daughters minus the category bound in the branching.}
\cite{Gunji91}}. This treatment is similar to the one introduced by
\cite{HPSG} for the treatment of English topicalization. However,
Japanese topicalization is fundamentally different from English
one. Firstly, it occurs more frequently. Up to 50\% of the sentences
are concerned (\cite{Yoshimoto97}). Secondly, there are examples where
the topic occurs in the middle of the sentence, unlike the English
topics that occur sentence-initially. Thirdly, suppressing of verbal
arguments in Japanese could be called more a rule than an exception in spoken
language. The SLASH approach would introduce traces in almost every
sentence. This, in connection with scrambling and suppressed particles,
could not be restricted in a reasonable way. If one follows Gunji's
interpretation of those cases, where the topic-NP can be interpreted as
a noun modifying phrase, a genitive gap has to be assumed. But this
leads to assuming a genitive gap for every NP that is not
modified. Further, genitive modification can be iterated.  Finally, two
or three occurences of NP-{\em wa} are possible in one utterance.
Thus, we decided to assign topicalized sentences the same syntactic
structure as non-topicalized sentences and to resolve the problem on
the lexical level.
The topic particle is, on the syntactic level, interpreted as a verbal
adjunct. The binding to verbal arguments is left to the semantic
interpretation module in VERBMOBIL, see figure~\ref{figure:topic-p-avm}.

{\em mo} is similar to {\em wa} in some aspects.  It can mark a
predicative adjunct and can follow {\em de} and {\em ni}. But it can
also follow {\em wa}, an adjective and a sentence with question mark:

\vspace{-0.3cm}
\enumsentence{
\label{eq:103}

\JPChar{dekiru\\can}
\JPChar{ka\\QUE}
\JPChar{mo\\MO}
\JPChar{shiremaseN\\do not know}\\
(I don't know if I can)
}

\vspace{-0.3cm}
{\em mo} is a particle that has the head of a
topic-adverbial particle, but a different subcategorization frame than
{\em wa}.
{\em koso} is another topic particle that can occur after nouns,
postpositions or adverbial particles.

\subsection{Omitted Particles}

Some particles can be omitted in Japanese spoken language. Here is an
example from the {\VM} corpus:

\vspace{-0.3cm}
\enumsentence{

\label{eq:89}

\JPChar{rokugatsu\\June}
\JPChar{\O\\\O}
\JPChar{juusaNnichi\\13th}
\JPChar{no\\NO}
\JPChar{kayoubi\\Tuesday}
\JPChar{\O\\\O}
\JPChar{gogo\\afternoon}
\JPChar{kara\\KARA}
\JPChar{wa\\WA}
\JPChar{ikaga\\good}
\JPChar{deshou\\COP}
\JPChar{ka\\QUE}\\
(Would the 13th of June suit you?)
}

\vspace{-0.3cm}
This phenomenon can be found frequently in connection with pronouns and
temporal expressions in the domain of appointment
scheduling. \cite{Hinds77} assumes that exclusively {\em wa} can be
suppressed. \cite{Yatabe93} however shows that there are contexts,
where {\em ga}, {\em wo} or even {\em e} can be omitted. He assigns it
as `phonological deletion'. \cite{Kuroda92}
analyses omitted {\em wo} particles and explains these with
linearization: a particle {\em wo} can only be omitted, when it occurs directly
before a verb. \cite{Yatabe93} however gives examples to prove the
opposite. 
It can be observed that NPs without particles
can fulfill the functions of a verbal argument or of
a verbal adjunct (ex.~\ref{eq:89}). We
decided to interpret these NPs as verbal adjuncts and to leave the
binding to argument positions to the semantic interpretation.
NPs thus get a MOD value that allows them to modify nonauxiliary
verbs.

\subsection{{\em ga}-Adjuncts}
\label{ga-adjuncts-sect}

One can find several examples with {\em ga} marked adjuncts in the
{\VM} data. On the level of information structure it is said that {\em
ga} marks neutral descriptions or exhaustive descriptions
(c.f. \cite{Gunji87}, \cite{Kuno73}).  Gunji analyses these exhaustive
descriptions syntactically in the same way as he analyses his `type-I
topicalization'. They build adjuncts that control gaps or reflexives in
the sentence. He views {\em ga} marked adjuncts without control
relations as relying on a very specialized context.  However, his
treatment leads to problems.  Firstly, in all cases, where {\em ga}
marks a constituent that is subcategorized as {\em ga}-marked by the
verb, a second reading is analysed that contains a {\em ga} marked
adjunct controlling a gap. This is not reasonable. The treatment of the
different meaning of {\em ga} marking and {\em wa} marking belongs to
the semantics and not into the phrase structure. Secondly, this
treatment assumes gaps. We already criticized this in connection with
topicalization.  Therefore, we do not need reflexive control at the
moment. However, it contains mostly examples with {\em ga} marked
adjuncts without syntactic control relation to the rest of the
sentences.

At the level of syntax, we do not decide whether a {\em ga}-marked
subject or object is a neutral description or an exhaustive
listing. This decision must be based on context information, where it
can be ascertained whether the noun phrase is generic, anaphoric or
new. We distinguish occurrences of NP+{\em ga} that are verbal
arguments from those that are  adjuncts.
The examples for {\em ga}-marked adjuncts in the {\VM} dialogues
either describe a temporal entity or a human.
All cases found are predicate modifying. To further restrict 
exhaustive interpretations, we introduced selectional
restrictions for the marked NP, based on observations in the data.

\begin{centering}\section{Conclusion}\end{centering}

The syntactic behaviour of Japanese particles has been analysed based on
the {\VM} dialogue data. We observed 25 different particles in 800
dialogues on appointment scheduling. It has been possible to set up a
type hierarchy of Japanese particles. We have therefore adopted a
lexical treatment instead of a syntactic treatment based on phrase
structure. This is based on the different kinds of modification and
subcategorization that occur with the particles. We analysed the
Japanese particles according to their cooccurrence potential,
their modificational behaviour and their occurrence in verbal
arguments.

We clarified the
question which common characteristics and differences between the
individual particles exist. A classification in categories was carried
out. After that a model hierarchy could be set up for an HPSG
grammar. The simple distinction into case particles and postpositions
was proved to be insufficient. The assignment of the grammatical
function is done by the verbal valency and not directly by the case
particles. The topic particle is ambiguous. Its binding is done by
ambiguity and underspecification in the lexicon and not by the
Head-Filler Rule as in the HPSG for English (\cite{HPSG}). 

The approach presented here is part of the syntactic analysis of
Japanese in the {\VM} machine translation system. It is implemented
in the PAGE parsing system \cite{Uszkoreit+al94}.  It has been proved to
be essential for the processing of a large amount of Japanese dialogue
data.

Further research concerning coordinating particles ({\em
to, ya, toka, yara, ka} etc.) and sentence end particles ({\em ka,
node, yo, ne} etc.) is necessary.

\begin{small}
\bibliography{/home/cl-home/siegel/literatur/lit}

\begin{thebibliography}{10}

\bibitem{Gunji87}
Takao Gunji.
\newblock {\em {J}apanese Phrase Structure Grammar}.
\newblock Dordrecht: Reidel., 1987.

\bibitem{Gunji91}
Takao Gunji.
\newblock An overview of {J}{P}{S}{G}: A constraint-based descriptive theory
  for {Japanese}.
\newblock In {\em Proceedings of {J}apanese Syntactic Processing Workshop}.
  Duke University, 1991.

\bibitem{Hinds77}
John Hinds.
\newblock Particle deletion in {J}apanese and {K}orean.
\newblock {\em Linguistic Inquiry}, 8(4):602--604, 1977.

\bibitem{Kuno73}
Susumo Kuno.
\newblock {\em The Structure of {J}apanese Language.}
\newblock Cambridge, Mass.: MIT Press., 1973.

\bibitem{Kuroda92}
S.-Y. Kuroda.
\newblock {\em {J}apanese Syntax and Semantics. Collected Papers.}, volume~22
  of {\em Studies in Natural Language and Linguistic Theory}.
\newblock Dordrecht: Kluwer Academic Publishers, 1992.

\bibitem{Miyagawa86}
Shigeru Miyagawa.
\newblock Predication and numeral quantifiers.
\newblock In William~J. Poser, editor, {\em Papers from the Second
  International Workshop on {J}apanese Syntax}, pages 157--191. CSLI, 1986.

\bibitem{Nightingale96}
Stephen Nightingale.
\newblock {\em An HPSG Account of the {J}apanese Copula and Related Phenomena}.
\newblock PhD thesis, University of Edinburgh, 1996.

\bibitem{Ono94}
Kiyoharu Ono.
\newblock Annularity in the distribution of the case particles {\em ga}, {\em
  o} and {\em ni} in {Japanese}.
\newblock {\em Theoretical Linguistics}, 20(1):71--93, 1994.

\bibitem{HPSG}
C.~Pollard and I.A. Sag.
\newblock {\em Head-Driven Phrase Structure Grammar}.
\newblock Chicago: University of Chicago Press., 1994.

\bibitem{Sadakane/Koizumi95}
Kumi Sadakane and Masatoshi Koizumi.
\newblock On the nature of the "dative" particle {\em ni} in {Japanese}.
\newblock {\em Linguistics}, 33:5--33, 1995.

\bibitem{Tsuda/Harada96}
Hiroshi Tsuda and Yasunari Harada.
\newblock Semantics and pragmatics of adnominal particle no in {Quixote}.
\newblock In Takao Gunji, editor, {\em Studies in the Universality of
  Constraint-Based Structure Grammars.} Osaka., 1996.

\bibitem{Tsujimura96}
Natsuko Tsujimura.
\newblock {\em An Introduction to {J}apanese Linguistics}.
\newblock Blackwell, Cambridge, 1996.

\bibitem{Uszkoreit+al94}
Hans Uszkoreit, Rolf Backofen, Stephan Busemann, Abdel~Kader Diagne,
  Elizabeth~A. Hinkelman, Walter Kasper, Bernd Kiefer, Hans-Ulrich Krieger,
  Klaus Netter, G{\"u}nter Neumann, Stephan Oepen, and Stephen~P. Spackman.
\newblock {DISCO}---an {HPSG}-based {NLP} system and its application for
  appointment scheduling.
\newblock In {\em Proceedings of COLING-94}, pages 436--440, 1994.

\bibitem{Yatabe93}
Shoichi Yatabe.
\newblock {\em Scrambling and {J}apanese Phrase Structure}.
\newblock PhD thesis, Stanford University., 1993.

\bibitem{Yoshimoto97}
Kei Yoshimoto.
\newblock {\em Tense and Aspect in {J}apanese and {E}nglish}.
\newblock PhD thesis, Universit{\"a}t Stuttgart, 1997.

\end{thebibliography}
\end{small}

\end{document}